\documentclass{article}

\usepackage[nonatbib,final]{nips_2017}

\usepackage[utf8]{inputenc} 
\usepackage[T1]{fontenc}    
\usepackage{hyperref}       
\usepackage{url}            
\usepackage{booktabs}       
\usepackage{amsfonts}       
\usepackage{nicefrac}       
\usepackage{microtype}      
\usepackage{enumitem}
\usepackage{amsthm}
 
\theoremstyle{definition}
\newtheorem{definition}{Definition}[section]

\usepackage[symbol*]{footmisc}
\DefineFNsymbolsTM{myfnsymbols}{
  \textasteriskcentered *
  \textdagger    \dagger
  \textdaggerdbl \ddagger
  \textsection   \mathsection
  \textbardbl    \|%
  \textparagraph \mathparagraph
}%
\setfnsymbol{myfnsymbols}

\usepackage{amssymb}
\usepackage{amsmath}
\usepackage{bm}
\newcommand\Tstrut{\rule{0pt}{2.6ex}}         
\newcommand\Bstrut{\rule[-0.9ex]{0pt}{0pt}}   
\usepackage{graphicx}
\usepackage{float}

\title{Survival-Supervised Topic Modeling with Anchor Words:
Characterizing Pancreatitis Outcomes}

\author{
  George H.~Chen \\
  Carnegie Mellon University \\
  \And
  Jeremy C.~Weiss \\
  Carnegie Mellon University \\
}

\begin{document}

\maketitle

\vspace{-1.5em}
\begin{abstract}
\vspace{-1em}
We introduce a new approach for topic modeling that is supervised by survival
analysis. Specifically, we build on recent work on unsupervised topic modeling
with so-called anchor words by providing supervision through an elastic-net
regularized Cox proportional hazards model. In short, an anchor word being
present in a document provides strong indication that the document is
partially about a specific topic. For example, by seeing ``gallstones'' in a
document, we are fairly certain that the document is partially about medicine.
Our proposed method alternates between learning a topic model and learning a
survival model to find a local minimum of
a block convex optimization problem. We apply our proposed
approach to predicting how long patients with pancreatitis admitted to an intensive care unit
(ICU) will stay in the ICU.
Our approach is as accurate as the best of a variety of baselines
while being more interpretable than any of the baselines.
\end{abstract}

\vspace{-1em}
\section{Introduction}
\label{sec:intro}
\vspace{-0.5em}

Health care is replete with problems of understanding how much time there is
before a critical event happens, such as how long a patient has to live or
how long a patient will stay in a hospital intensive care unit (ICU). These
time durations can be modeled by numerous survival analysis techniques.
Especially as survival analysis models learned
can inform costly interventions, ensuring that they are clinically
interpretable is essential.
At the same time,
these models now have to cope with an enormous
number of measurements collected per patient in electronic health records
for which we do not fully understand how all the measurements
relate. In this paper, we aim to address the twin objectives of learning
how
measurements relate in the form of a topic model, and learning how
the topics can make survival predictions at the patient level.

%

We phrase the problem setup 
in
terms of predicting how long pancreatitis patients admitted to an
ICU will stay in the ICU. We have training data of $n$
patients.
For each patient, we know how many times each of $d$ words
appears. The dictionary of words is pre-specified based on clinical events derived from electronic health record data.
We denote $X_{w,i}$ to be the
number of times word $w \in \{1,\dots,d\}$ appears for
patient $i \in \{1,\dots,n\}$. Viewing $X$ as a $d$-by-$n$ matrix,
the $i$-th column of $X$ can be thought of as
the feature vector for the $i$-th patient. As for the training label for
the $i$-th patient, we have two recordings: $R_i \in \{0,1\}$ which
indicates whether we know the ICU length of stay for the patient, and
$Y_i \in \mathbb{R}_+$ is the ICU length of stay
for the patient if $R_i=1$, or the ``censoring time'' if $R_i=0$. The
censoring time gives a lower bound on how long the $i$-th patient was in
the ICU.

Our goal is to discover
topics for the~$d$ words that help
predict how long a pancreatitis patient newly admitted into the ICU
will stay in the ICU.
Note that an unsupervised topic model would
not use any of the training labels, instead
learning topics using only the word count matrix~$X$. Meanwhile, in
survival analysis, a standard approach would learn a survival model
using all the patients' feature vectors and labels but would not
learn thematic structure in the different features, \emph{e.g.}, topics.

Our contributions are twofold:
\begin{itemize}[leftmargin=*,nolistsep,itemsep=0.25em]
\item 
To jointly learn both a topic and a survival model,
we extend a recently proposed unsupervised topic modeling approach using
so-called \textit{anchor words}
\cite{anchor_words,almost_anchor_words,fast_anchor_words}
to be supervised by an elastic-net regularized Cox
proportional hazards model \cite{glm_elastic}
(which generalizes the standard Cox proportional hazards model
\cite{coxph}).
Anchor words, to be described in detail shortly, act as exemplar words for
the topics. Our proposed method 
finds anchor words and then
alternates between learning a topic model and learning a survival model,
corresponding to finding a local minimum of a block convex optimization
problem. We remark that how we combine topic modeling and survival analysis is
similar to an existing approach \cite{survLDA} that combines Latent Dirichlet Allocation (LDA)
\cite{LDA} with the standard Cox proportional hazards model; however the
resulting algorithms are syntactically very different. We explain why
topic modeling with anchor words is preferable to LDA.
\item
We apply our proposed method to predicting how long pancreatitis
patients admitted to an ICU will stay in the ICU. Our method is as accurate
as the best of a variety of baseline methods yet comes with new
interpretative advantages. In particular, the best performing baselines
either do not learn how different
measurements relate, or do not learn clinically interpretable topics.
In contrast, our proposed
method simultaneously learns topics corresponding to different collections of
events likely to have happened to patients, how presence of a topic
contributes to lengthening or shortening a patient's stay in the ICU,
and what anchor words are for the topics. The anchor words found turn out
to correspond to catheters used
by physicians. Although our method is only competitive with
and does not necessarily outperform the best of the baselines considered, we believe
that among methods with similar performance, we should opt for the most
interpretable.
\end{itemize}


\vspace{-1em}
\section{Background: Topic Modeling with Anchor Words}
\vspace{-0.7em}

The intuition behind anchor words is that
some words are very
strong indicators of specific topics being in a document.
For example, for text documents,
if the word ``gallstones'' appears in a document, we can be fairly certain that
the document
is partially about medicine. Similarly, ``401k'' can be an indicator for
finance, and ``Oscar-winning'' for movies. These are examples of anchor words.
Importantly, not every document about medicine has to have the
anchor word ``gallstones'' in it. In fact, anchor words could occur
infrequently. In the context of characterizing outcomes of a
specific disease like pancreatitis, the words correspond to specific
measurements or events such as a patient's lab measurements quantized
to discrete values, whether a patient was previously
put on a specific
kind of intravenous fluid, etc. The topics consist of frequently co-occurring
measurements or events.

Anchor word topic modeling has
three properties that altogether make it preferable over a more
standard topic modeling approach like LDA. First, we can find anchor words quickly.
In particular, there is a fast,
provably correct algorithm that finds an accurate approximation of
the anchor words (see Algorithm~4 and Theorem~4.3 by~Arora \textit{et al.}~\cite{fast_anchor_words}).
Second, we can check whether the anchor words
have enough expressive power to explain the full dictionary of words. This
property results from a key interpretation of anchor words: anchor words
turn out to, in some sense, be proxies for their respective topics. Under the
assumption that every topic we aim to learn has an anchor word, then there is
a way to represent every word as a convex combination of the anchor words.
As a result, we can efficiently check whether a word is representable
by the anchor words found (the problem reduces to checking whether a
point is inside a convex hull). If not, then it means
that more anchor words are needed to explain the full dictionary, \emph{i.e.}, more
topics are needed under the assumption that every topic has an anchor word.
Alternatively it means that we can quickly identify words to remove from the
dictionary if we do not want to use more topics. Lastly, anchor word topic
modeling makes no assumption on the topics being uncorrelated, in contrast to
LDA.

\setlength{\abovedisplayskip}{0pt}
\setlength{\belowdisplayskip}{0pt}

To build up to the formal definition of an anchor word, we establish some
notation used in the rest of the paper. Topic models generally
assume that
each document (in our survival context, each document corresponds to a patient)
 contains a mixture of topics, and each topic has a distribution
over words. If we assume that there are $k$ topics,
then words in document $i\in\{1,\dots,n\}$ are assumed to be sampled
independently from the following distribution over words $w\in\{1,\dots,d\}$:
\begin{align*}
M_{w,i} & \triangleq\mathbb{P}(\text{word }w\mid\text{document }i) 
    =\sum_{g=1}^{k}\underbrace{\mathbb{P}(\text{word }w\mid\text{topic }g)}_{\triangleq A_{w,g}}\underbrace{\mathbb{P}(\text{topic }g\mid\text{document }i)}_{\triangleq W_{g,i}}.
\end{align*}
As the notation suggests, we can view topic
modeling in terms of nonnegative matrices $M\in\mathbb{R}_{+}^{d\times n}$,
$A\in\mathbb{R}_{+}^{d\times k}$, and $W\in\mathbb{R}_{+}^{k\times n}$.
The $i$-th column of $M$ is document $i$'s word distribution. The
$g$-th column of $A$ is topic $g$'s word distribution. The $i$-th
column of $W$ is document $i$'s topic distribution. Then the
equation above can be written as the nonnegative matrix
factorization~${M=AW}$. We do not observe $M$. What we
instead have is
the word count matrix $X\in\mathbb{R}_{+}^{d\times n}$
described in Section~\ref{sec:intro},
where $X_{w,i}$ is the number of times word~$w$ appears in document
$i$.
Denoting $m_i$ to be the number of words in the $i$-th document, then
$X_{w,i} \sim \text{Binomial}(m_i, M_{w,i})$. Different topic models
place different assumptions on matrices $A$ and $W$.

\begin{definition}
An \textit{anchor word} for a topic is defined as a word that has positive
probability (within word-topic matrix $A$) of
appearing only for that one specific topic.
(For example, if
topic~3 has word~5 as an anchor word, then in the
row~5 of $A$, the only nonzero entry appears in column~3.)
\end{definition}

\vspace{-.6em}
Anchor word topic modeling learns word-topic matrix $A$ given word
count matrix $X$ under the assumption that $A$ is deterministic and unknown,
and
columns of $W$ are
stochastically generated.
The key assumption used in learning an anchor word topic model is that
every topic that we learn has an anchor word. Supposing that there are
$k$ topics, then every topic $g \in \{1,\dots,k\}$ has an associated
anchor word $a_g \in \{1,\dots,d\}$ (if a topic has multiple anchor
words, we can arbitrarily choose one of them to use). 
Anchor word topic model learning proceeds by first finding
what the anchor words are given word count matrix $X$ using a greedy
(albeit provably accurate)
combinatorial search. Next, we
determine how to represent every word in the dictionary in terms of
the anchor words, which amounts to finding how to represent each word
as a specific convex combination of the anchor words. Lastly, a
Bayes' rule calculation can be used to estimate $A$. For details, see
the paper by Arora \textit{et al.}~\cite{fast_anchor_words}.

\vspace{-0.7em}
\section{Method}
\vspace{-0.7em}



\setlength{\abovedisplayskip}{4pt}
\setlength{\belowdisplayskip}{4pt}

Following how Dawson and Kendziorski combine LDA with the standard Cox
proportional hazards model to form the \textsc{survLDA} model
\cite{survLDA}, we set the $i$-th training
patient's feature vector in the Cox proportional hazards model to be
the estimated probabilities of each of the $k$ topics appearing in the
$i$-th patient/document. Put another way, this is a layered composition:
the input word count matrix $X$ goes through an anchor word topic modeling
layer that outputs each patient's topic proportions. These topic
proportions across all patients are then treated as a single input to an
elastic-net regularized Cox proportional hazards model layer that
outputs the labels ($R_i$'s and $Y_i$'s for whether censoring happened and the survival/censoring
times).\footnote{Despite this layered composition explanation, we remark
that anchor word topic modeling does not readily fit into the
framework of standard black box inference
(\emph{e.g.}, autoencoding variational Bayes~\cite{AEB,rezende_et_al_2014}) approaches
due to its multiple-stage learning involving first solving a combinatorial
optimization greedily. Separately, black box inference struggles to learn 
topic models; only recently has there been a method~\cite{prodLDA} developed that uses black box inference to learn LDA with
carefully chosen
optimization parameters and a Laplace approximation.
Topic models compatible with black box inference
are not yet as theoretically well-understood as
anchor word topic modeling.
Meanwhile \textsc{survLDA} uses standard (non-black-box) variational EM
\cite{survLDA}.}

We now state our proposed algorithm. Given a pre-specified number of
topics $k$, we first compute an estimate
$\widehat{a}_1,\dots,\widehat{a}_k$ of the anchor words
using the same algorithm as in unsupervised anchor word topic modeling.\footnote{%
We run Algorithm~4 by~Arora \textit{et al.}~\cite{fast_anchor_words}. Since this algorithm is randomized, to make the result more stable (in fact, seemingly deterministic), we run it a pre-specified number of times, and pick whichever specific run has anchor words that have appeared the most across all the runs. Details are in the longer version of this paper.
}
To describe the optimization problem to be numerically optimized next,
we introduce some variables that we can compute based on the word
count matrix $X$. First, the $i$-th column of $X$ corresponds to the
$i$-th patient's word counts. Let $\overline{X}\!~^{(i)} \in [0,1]^d$ denote the
$i$-th patient's word counts normalized to sum to~1. From $X$, we also compute the
$d$-by-$d$ word co-occurrence matrix~$Q$.\footnote{%
Technical details for how to count word self co-occurrences and
how to account for varying document lengths in forming $Q$ are in
the supplementary material Section 4.1~of Arora \textit{et al.}~\cite{fast_anchor_words}.
} Let $\overline{Q}_w\in[0,1]^d$ denote the $w$-th row of
$Q$ normalized to sum to 1.
Let $\theta_{w,g} = \mathbb{P}(\text{topic }g\mid\text{word }w)$ and
$\beta \in \mathbb{R}^k$ be Cox regression coefficients.
We numerically solve: 
\begin{align*}
\underset{\substack{\theta\in\mathbb{R}_+^{d\times k}, \\
                    \beta\in\mathbb{R}^k}}{\text{minimize:}}\;
& \!\!\!\!\sum_{\substack{w=1 \\
        \text{ s.t.~}w\notin\{\widehat{a}_1,\dots,\widehat{a}_k\}}}^d\!\!\!\!\!\!\!\!
    D_{\text{KL}}\big(
      \overline{Q}_w\big\|{\textstyle \sum_{g=1}^k}\theta_{w,g}\overline{Q}_{\widehat{a}_g}\big) 
      +\sum_{i=1}^{n}
  R_i
  \Big(-\beta^{T} (\overline{X}\!~^{(i)})^T \theta
       +\log\!\!\!\!\!\!
          \sum_{\substack{j=1 \\
                          \text{s.t.~}Y_j \ge Y_i}}^{n}
            \!\!\!\!\!\!
            \exp(\beta^{T}(\overline{X}\!~^{(j)})^T \theta)\Big) \nonumber \\
 & \!\!\!\qquad+\lambda\big(\alpha\|\beta\|_{1}+{\textstyle \frac{1}{2}}(1-\alpha)\|\beta\|_{2}^{2}\big) \nonumber \\
\text{subject to:}\; & {\textstyle \sum_{g=1}^{k}}\theta_{w,g}=1\text{ for each }w\in\{1,\dots,d\},\;\;\,\,
\theta_{\widehat{a}_g, h} = \mathbf{1}\{g = h\}\text{ for each }g,h\in\{1,\dots,k\},
\label{eq:opt-anchor-survival}
\end{align*}
where $\mathbf{1}\{\cdot\}$ is the indicator function that is 1 when
its argument holds and 0 otherwise, and constants $\lambda>0$ and $\alpha>0$
are hyperparameters controlling the elastic-net regularization on Cox
regression coefficients $\beta$.
The first term in the objective function finds a representation of each
word $\overline{Q}_w$ as a convex combination $\theta_w$ of the anchor words
$\overline{Q}_{\widehat{a}_1}, \dots, \overline{Q}_{\widehat{a}_k}$.
The second and third terms in the
objective function correspond to the elastic-net regularized Cox
partial log likelihood, where the $i$-th patient's feature vector is set to the
patient's estimated proportion of words coming from each topic
$(\overline{X}\!~^{(i)})^T \theta \in [0,1]^k$. Removal of these two
survival-related terms from
the objective function results in unsupervised anchor word topic modeling
(word-topic
matrix $A$ can be recovered from $\theta$'s via Bayes' rule).
 By fixing $\beta$ and
optimizing $\theta$ and vice versa, we alternate between solving two
convex programs that learn a topic model (updating
$\theta$) 
 and learn a
survival model (updating $\beta$). 
\vspace{-1em}
\section{Experimental Results}
\vspace{-1em}

\begin{figure}[b]
\vspace{-1.5em}
\centering
\tiny
\begin{sc}
\begin{tabular}{lccc}
\hline
Method & RMSE (days) & MAE (days) & c-index \Tstrut\Bstrut\\
\hline
KM & 14.4 & 7.4 & --- \Tstrut\\
SVD Cox & 14.4 & 7.2 & 0.59 \\
SVD Aalen & 13.6 & 7.4 & 0.59 \\
SVD AFT & 13.5 & 6.9 & 0.59 \\
Lasso Cox & 13.5 & \textbf{6.8} & \textbf{0.63} \\
EN Cox & 13.5 & \textbf{6.8} & \textbf{0.63} \\
USAW & 14.3 & 7.3 & 0.59 \\
SAW & \textbf{13.3} & \textbf{6.8} & 0.59 \Bstrut\\
\hline
\end{tabular}
~~~~
\begin{tabular}{lc}
\hline
Anchor word of topic & $\beta$ 
 \Tstrut\Bstrut\\
\hline
Arterial line insertion & -4.5 \Tstrut\\ 
\quad 18 gauge insertion & \\ 
\quad Transfer from hospital & \\ 
\quad Private insurance & \\ 
\quad Male & \\ 
Multi lumen insertion & 2.3 \\ 
\quad 20 gauge insertion & \\ 
\quad White & \\ 
\quad Female & \\ 
\quad Emergency room admission & \\ 
PICC line insertion & 0.4 \\ 
\quad 20 gauge insertion & \\ 
\quad Emergency room admission & \\ 
\quad Male & \\ 
\quad Hospital admit date & \\ 
\hline
\end{tabular}
\end{sc}
\caption{(Left) Test set ICU length of stay RMSE and MAE, and c-index by method. (Right) A few top words in topics with nonzero $\beta$ coefficients.} 
\vspace{-3em}
\label{rmses}
\end{figure}

For our experiments, we looked at patients with pancreatitis in the MIMIC III dataset 
\cite{johnson2016mimic} who required admission to the ICU, amounting to a total of 371 individuals.
Features extracted include demographics, medications, billing codes, procedures, laboratory measurements, events recorded into charts, and vitals. Each record was placed into a 4-column format
\emph{patient id}, \emph{time}, \emph{event}, and \emph{event value}.
Preprocessing details such as how words are chosen are given in the longer version of this
paper. At a high-level, we use patient data up until they first enter ICU; any data afterward is
not present in the word counts $X$. Words are derived from pairs (\textit{event}, \textit{event value}), with the event value discretized if it is continuous.
Words too similar in frequency across different ICU lengths of stay are filtered out.
We split the data into 75\% training and 25\% test. For the methods that we
compare, 3-fold cross-validation within the training set is used to select the
best
algorithm parameters as to minimize
root-mean-square error (RMSE) in predicted median ICU length of stay compared to the true
length of stay. After finding these parameters we train each method on the full training set.
We compared our proposed supervised anchor word topic modeling method (abbreviated ``SAW'' in Figure~\ref{rmses}) with a two-stage approach that learns an unsupervised anchor word topic model first and then learns an elastic-net regularized Cox model without joint optimization (``USAW''). We also compare against survival analysis baselines that do not use anchor words: Kaplan Meier (``KM''), Cox proportional hazards (``Cox''), Aalen additive models (``Aalen''), and accelerated failure time models with a Weibull distribution (``AFT''). Because our task has many more features than patients ($d \gg n$), we used singular value decomposition (SVD) as a preprocessing step to produce a low-dimensional feature space for the last three methods (we remark that this low-dimensional representation can be thought of as a topic model that is \textit{not} clinically interpretable).  Separately for the Cox proportional hazards model, we also compare with both its lasso-regularized (``Lasso Cox'') and, more generally, elastic-net regularized (``EN Cox'') variants, both of which are clinically interpretable but do not learn topics. 
We were unable to acquire the \textsc{survLDA} code of Dawson and Kendziorski.
We report test set RMSE's, mean absolute errors (MAE's), and c-indices in
Figure~\ref{rmses} (left), and anchor words (found by our proposed method SAW) for all three topics with nonzero
$\beta$ coefficient (out of a total of $k=5$ topics; as with other parameters, $k$ was selected via cross-validation) along with a few of each topic's most frequently occurring words in Figure~\ref{rmses} (right).

The anchor words focus on catheters physicians used, with the ones associated with more severe illness, \emph{e.g.}, arterial lines and lower-gauge catheters, being part of the topic with a negative $\beta$ coefficient, indicative of longer ICU stays. The five topics learned roughly typify: severe, non-severe, demographic indicators, laboratory indicators, and heart-related concepts. While these last two topics are interpretable, they were not deemed (sufficiently) predictive with $\beta=0$ coefficients. 
In terms of prediction accuracy metrics, our proposed method is competitive with (although not necessarily better than) the best-performing baselines (SVD AFT, Lasso Cox, EN Cox).

With respect to interpretability, the results illustrate that SAW identifies clusters of outcome-relevant, clinical events and provides a level of risk associated with those clusters. Among thousands of highly co-linear events, SAW produces a digestible output that both matches clinical knowledge, \emph{e.g.}, catheter size and transfer as indicators of severity, and offers hypotheses about unforeseen relationships and substitutability. Clinical concepts of organ failure and hemoconcentration are known predictors of pancreatitis severity and are closely related to the anchor words with nonzero~$\beta$. 

\bibliographystyle{plain}
\bibliography{as}

\end{document}